\newif\ifdraft\draftfalse
\documentclass[sigconf]{aamas} 
\usepackage{balance} % for balancing columns on the final page

\usepackage{xcolor}

\usepackage{stfloats}

\usepackage{soul}
\usepackage{url}
\usepackage[utf8]{inputenc}
\usepackage{graphicx}
\usepackage{amsmath,amsthm}

\usepackage{booktabs}
\usepackage{listings, array, blkarray}
\usepackage{bbm, dsfont}
\usepackage{multicol}
\usepackage{wrapfig,multirow}
\usepackage{algorithm}
\usepackage{algorithmic}

\usepackage[caption=false]{subfig}

\setcopyright{ifaamas}
\acmConference[AAMAS '21]{Proc.\@ of the 20th International Conference on Autonomous Agents and Multiagent Systems (AAMAS 2021)}{May 3--7, 2021}{Online}{U.~Endriss, A.~Now\'{e}, F.~Dignum, A.~Lomuscio (eds.)}
\copyrightyear{2021}
\acmYear{2021}
\acmDOI{}
\acmPrice{}
\acmISBN{}

\acmSubmissionID{???}

\title[RL for Active Screening]{Active Screening for Recurrent Diseases: \\A Reinforcement Learning Approach}

\author{Han-Ching Ou, Haipeng Chen, Shahin Jabbari and Milind Tambe}
\affiliation{%
  \institution{Harvard University}
  \city{Cambridge} 
  \state{Massachusetts} 
  \postcode{02138}
}
\email{{hou@g., hpchen@seas., jabbari@seas., milind_tambe@}harvard.edu}

\begin{abstract}
Active screening is a common approach in controlling the spread of recurring infectious diseases such as tuberculosis and influenza. In this approach,  health workers periodically select a subset of population for screening. However, given the limited number of health workers, only a small subset of the population 
can be visited in any given time period. Given the recurrent nature of the disease and rapid spreading, the goal is to minimize the number of infections over a long time horizon. 
Active screening can be formalized as a sequential combinatorial optimization over the network of people and their connections. 
The main computational challenges in this formalization arise from i) the combinatorial nature of the problem, ii) the need of sequential planning and iii) the uncertainties in the infectiousness states of the population.

Previous works on active screening fail to scale to large time horizon while fully considering the future effect of current interventions. In this paper, we propose a novel reinforcement learning (RL) approach based on Deep Q-Networks (DQN), with several innovative adaptations that are designed to address the above challenges. First, we use graph convolutional networks (GCNs) to represent the Q-function that exploit the node correlations of the underlying contact network. Second, to avoid solving a combinatorial optimization problem in each time period, we decompose the node set selection as a sub-sequence of decisions, and further design a two-level RL framework that solves the problem in a hierarchical way. 
Finally, to speed-up the slow convergence of RL which arises from reward sparseness, we incorporate ideas from curriculum learning into our hierarchical RL approach. We evaluate our RL algorithm on several real-world networks. Results show that our RL algorithm can scale up to 10 times the problem size of state-of-the-art (the variant that considers the effect of future interventions but un-scalable) in terms of planning time horizon. Meanwhile, it outperforms state-of-the-art (the variant that scales up but does not consider the effect of future interventions) by up to $33\%$ in solution quality.
\end{abstract}


\newcommand{\BibTeX}{\rm B\kern-.05em{\sc i\kern-.025em b}\kern-.08em\TeX}

\begin{document}

%%% The following commands remove the headers in your paper. For final 
%%% papers, these will be inserted during the pagination process.

\pagestyle{fancy}
\fancyhead{}

%%% The next command prints the information defined in the preamble.

\maketitle 
%\thispagestyle{plain}
%\pagestyle{plain}

%%%%%%%%%%%%%%%%%%%%%%%%%%%%%%%%%%%%%%%%%%%%%%%%%%%%%%%%%%%%%%%%%%%%%%%%

\section{Introduction}

Active screening (or contact tracing) aims at selecting a subset of nodes in a social network for screening, so as to prevent the spread of transmissive diseases.
It is a system used by health workers globally to stop infectious diseases, including influenza, sexually transmitted diseases, and tuberculosis.
When an individual is tested positive, they are marked as infected. The health worker, or contact tracer, records who else have been exposed and marks them as contacts or potentially infected individuals. 
%Such contact could be a certain kind of physical contact or simply sharing the same space with the infected one for a long enough time like flights, depending on how infectious the targeted disease is. \sj{remove this line} 
%The infected person may or may not have symptoms when the contact happens. Moreover, the contacted person may not have any symptom but being infectious by the time the infected person was tested. \sj{what does this mean? are you talking about incubation?} 
The potentially infected individuals might not voluntarily seek treatment and testing. In case any of such individuals are infected, they can spread the disease further. Active screening aims to target these individuals and slow down the spread of the disease~\cite{eames2003contact,taylor1994chlamydia}.%\sj{some citation to epi literature would be useful}.

There are many challenges in implementing active screening~\cite{ou2020}. First, not every individual in a social network can be screened due to limited resources (in this case contact tracers or amount of tests available). Therefore, for each screening round, we need to optimally select a subset of nodes which is a \textit{combinatorial optimization} problem. Second, the health states of the many individuals in the network are unknown (sans people who get tested actively or passively). Finally, the above challenges are significantly amplified when \textit{sequential planning} is involved as we need to account for the future effects of current screening actions.

%However, we show that solving the optimal multi-round active screen problem is extremely challenging. First, the exact state of each individual (i.e., whether they are infected or susceptible) is \textit{uncertain} unless they self report or we do go and screen them %\sj{simply say ``screen them"}. %\sj{how can an exposed person self-report?}

Most of the computer science literature on infectious disease focuses on conditions leading to disease eradication, which often relies on the assumptions such as complete knowledge of individual health states~\cite{eames2003contact,ferretti2020quantifying}. %\sj{this is inaccurate. the disease can eradicate naturally. do they really need unlimited budget? why is the problem interesting with unlimited budget?}%Bridging the gap between the problem under ideal case with its real-world application is thus a crucial challenge.
Unfortunately, such knowledge of health states is not always available, making these approaches difficult to apply in many real-world scenarios.
%It is therefore critical to bridge the gap between methodologies designed for those simplistic settings and approaches that apply to real-world scenarios.
%Such a gap was not addressed before~\cite{ou2020} that models the active screening process as \textit{Act problem}. 
Recently~\citet{ou2020} address the challenges of uncertain states and limited budget. However, their approaches either do not scale with the planning time horizon or fail to fully account for future actions. See Section~\ref{sec:related} for more details and Section~\ref{sec:exp} for comparisons with our approach. %\sj{need to describe \citet{ou2020} in the related work if we have not done so. I also removed the next paragraph-- seems like we are using RL just because of its success and not really applicability to our problem}
Due to the superior performance of RL approaches in solving long term planning problems~\cite{mnih2015human,silver2016mastering,silver2017mastering}, in this paper we propose a novel RL approach that builds upon a powerful variant of RL called DQN~\cite{mnih2013playing}. We first formulate the multi-round active screening problem as a Markov Decision Process (MDP), where the state is a vector representing the probability of each node in the network being infected, and the action is to select which subset of nodes to actively screen. 
Due to the extremely high-dimensional state and action spaces, vanilla DQN algorithms cannot be directly applied to solve our problem efficiently. We therefore design several innovative adaptations over vanilla DQN that fully exploit the problem structure of multi-round active screening. First, we show that the node features in the underlying contact network are inter-correlated. To efficiently capture the intrinsic correlations between different nodes, we use GCNs as the function approximator to represent the Q-function. Second, because in each time period we need to select a subset of nodes to actively screen, this leaves vanilla DQN un-scalable as it needs to solve a combinatorial optimization problem in the action selection procedure. To avoid this we decompose the node set selection problem in each time period as a sub-sequence of decisions, and then design a novel two-level RL framework that solves the problem in a hierarchical manner. It has two types of agents. The primary agent works at the main sequence level and interacts with the environment, while multiple secondary agents work at the sub-sequence level and are responsible for generating actions sequentially within each time period. Last, we find that the reward signals for the secondary agents are sparse. To speed up the slow convergence of secondary agents' policies that arises from the sparseness of rewards, we incorporate ideas from curriculum learning into our algorithm. Intuitively, the algorithm warm-starts at the beginning of training with a simpler task, which has limited action choice and true state information. As the training goes on, the algorithm gradually increases task difficulty by providing uncertain state information and more action choice until the problem becomes the same as the original active screening problem.

Our main contributions are summarized as follows.
%\begin{itemize}
%\item[(1)] 
(i) We are the first to formulate the multi-round active screening problem for recurrent diseases as a Markov Decision Process (MDP). 
(ii) To solve the formulated MDP, we propose a novel solution algorithm on the basis of DQN, with several innovative adaptations that fully exploit the problem structure of the formulated MDP.
(iii) We conduct extensive experiments on various real-world networks with distinct network properties to evaluate the effectiveness of our proposed approach. The empirical results show that our approach can scale up to 10 times the problem size of state-of-the-art (the variant that considers the effect of future interventions but is not scalable) in terms of planning time horizon. Meanwhile, it outperforms state-of-the-art (the variant that scales up but does not consider the effect of future interventions) by up to $33\%$ in terms of the total number of healthy people. The robustness analysis shows that it works better than baselines even with network structure uncertainty. Interestingly, the policy analysis results show that compared with the baselines, our approach does not rely on node structural importance (e.g., degree and betweenness), and thus is fairer in the sense that it tends to spread the screening across different nodes. %\sj{what are some of the other insights?}
%\end{itemize}

\section{Related Work}
\label{sec:related}

%\sj{moved this from intro, should be discussed here: Previous studies of infectious diseases can be categorized as disease with permanent immunity~\cite{contact1,contact2,wangthesis,dava,ganesh2005effect} or without~\cite{drakopoulos2016network,drakopoulos2014efficient,scaman2016suppressing}. }\ho{It is mentioned in second and third para right?I thunk they are duplicated previously which is my bad.}
%Controlling the disease spread has always been an important field of study. 
Due to the close relationship between the spread of disease and the structure of contact networks, network epidemiology models have gained much more attention compared to 
%has caught people's attention as it better captures where and how the disease will spread than a 
homogeneous models~\cite{giesecke2017modern,liljeros2003sexual}. Numerous papers have studied the properties of these graph-based models. For example, epidemic threshold of the recurrent disease is a particular vibrant sub-discipline for recurrent disease~\cite{boguna2013nature,parshani2010epidemic,wang2003epidemic,pastor2001epidemic}. In these studies, the steady-state of the whole population is analyzed, including sufficient conditions for disease eradication.
%\hp{Suggest consistently using present tense for all discussion of related work.} 
A wide range of assumptions are used in these models, spanning from scale-free static graphs to random or even dynamic networks. %However, these analysis are all under assumptions of no additional human interventions are involved.
However, none of them consider human interventions.

Vaccination problem is another topic of great importance in the network epidemiology. For non-recurrent disease, when a vaccine exists, strategies should be developed to allocate the vaccine in the most efficient way. Due to the economic cost of vaccination and possible side effects, one may want to keep the number of vaccinated individuals small. These methods assume the intervention offers permanent immunity and thus focus on a one-shot action instead of multiple rounds of actions~\cite{holme2004efficient,contact1,contact2,zhang2015data,eigen1,eigen3,ren2018node}. %\sj{can we spread out the citations in multiple parts of this para besides the first sentence}\ho{Few of them have different assumption about graph like uncertainty, not sure we want to spend time talk about the difference and list separately}\sj{you want to cite all your claims in related work. The way it is written, it is unclear whether what you say after the first sentence is our ideas or other people work. If it is our ideas, it should not be listed in related work. If it is others, citation is needed. By spreading the citation you assure the reviewers that these are not just made up and we are referring to actual papers.}

For recurrent diseases, temporary quarantines, tracking or treatment are often considered as alternatives in the absence of permanent cures like vaccines. Furthermore, these interventions are required to be performed over several rounds given the recurrent nature of the disease. The multi-round intervention for network epidemiology is  first studied from a theoretical prospective~\shortcite{scaman2016suppressing,drakopoulos2016network,drakopoulos2014efficient}. 
They prove that the disease could be eradicated for a certain budget threshold in each round under the assumption that the infectiousness states of all the individuals in the network are perfectly observable. 
In reality, such a perfect observation is often not available. \citet{hoffmann2018cost} analyze the impact of state uncertainty, indicating even a small amount of uncertainty on state will have a large impact on the eradication time and the budget needed. \citet{ou2020} study the multi-round intervention with unknown states in the context of active screening or contact tracing. They prove the NP-hardness of the problem and provide a gradient based algorithm with two variants. The first variant takes future actions into account but scales poorly with the time horizon (number of rounds) and graph size. The second variant, which does not have the scalability issues of the first variant, does not account for future actions. However, for rapidly spreading diseases, planning for a long-term time horizon is an essential and important task. This is even more important for diseases that are hard to eliminate~\cite{maher2007planning}.
Since \citet{ou2020} is the closest work to us, we compare our results with them in details in Section~\ref{sec:exp}. 

RL has attracted a lot of interest from researchers in the machine learning and artificial intelligence communities~\cite{mnih2015human,silver2016mastering,silver2017mastering}. It is an experiment-driven and mathematical framework that trains an agent through trial and error~\cite{kaelbling1996reinforcement,sutton1998introduction,sutton2018reinforcement}. With the rise of deep learning, researchers further overcome the computational limitations of traditional RL by utilizing the representation power of deep neural networks~\cite{mnih2013playing,arulkumaran2017deep}, such as recurrent neural networks (RNNs)~ \cite{zaremba2014recurrent}, convolutional neural networks (CNNs)~\cite{krizhevsky2012imagenet} and graph convolutional neural networks (GCNs)~\cite{kipf2016semi}. The early attempt of our work can be found in \cite{ouactive} where standard reinforcement learning is applied yet unable to surpass~\citet{ou2020}. In this work, a novel hierarchical reinforcement learning framework that utilize curriculum learning is proposed. Such approach significantly enhance the performance not only compared to~\cite{ouactive} but also the previous state of the art~\citet{ou2020}.

Active screening in each time period is a combinatorial optimization problem, an important branch of optimization with numerous practical applications~\cite{hartmann2005phase}. They are usually NP-hard and thus polynomial-time heuristic algorithms for finding approximate solutions are often used in practice~\cite{johnson1974approximation}.
Recent advances use deep neural networks to learn heuristics for \textit{one time period} graph combinatorial problems~\cite{khalil2017learning,bello2016neural,kamarthi2019learning}. They have shown promising results by combining RL with different graph embedding methods~\cite{dai2016discriminative,kipf2016semi}. For multi-round problems, \citet{song2019solving} use a weight sharing technique that addresses problems with relatively small selection budget. To the best of our knowledge, no previous work in this thread of study has been applied to tackle active screening.

\section{Problem Formulation}
In this work, we focus on a sequential decision making problem with a large time horizon, where in each round (time step) we aim to optimally select which individuals in a social network to actively screen, so that the expected number of un-infected individuals over the time horizon is maximized. 
%\sj{using RL is not part of the problem description, this should move to solution}
\subsection{Problem Background}
The environment we consider is based on the well-known network (Susceptible-Infected-Susceptible) SIS model.
%\sj{no need to refer to related work, this is ``our" problem formulation and we discussed related work already}
SIS models capture the dynamics of recurrent diseases, where permanent immunity is not possible. 
%\sj{just a reminder here to motivate in intro why SIS is a reasonable model}
We adopt a discrete-time SIS model for modeling the disease dynamics propagating on a given graph\footnote{We use network and graph interchangeably to denote social networks.}, where each node represents an individual, and each edge indicates the link between individuals in which disease can spread.

\noindent \textbf{Disease model:} Given a graph $G = (V, E)$, each node $v\in V$ in the discrete-time SIS model can be in either of the two states, susceptible ($S$) or infected ($I$). In each time step (or round) $t$, similar to the independent cascade model~\cite{KempeKT03}, each node in $I$ state may infect its neighboring nodes that are in $S$ state with a fixed probability $\beta$. Each infected node may also be cured and become susceptible again with a fixed probability $\gamma$. 
This represents the probability of a node going for a passive screening, which means that the case is discovered while the individual voluntarily visits a health facility. We denote the true infectiousness state of a given node $v$ at time $t$ as $\mathbf{x}^{v}_{t}$. 
This true state is assumed to be unknown and the contact network is assumed to be known to the health workers.\footnote{We will also show empirically in Sec.~\ref{sec:exp} how our algorithm works when the contact network is unknown.}

\noindent \textbf{Intervention model: } We wish to automatically train an active screening agent to learn a policy that controls the spread of disease. The intervention we consider involves multi-rounds of active screening over a time horizon of $T$ days. %Specifically, we are allocating a limited number $k$ of health workers over a time horizon of $T$ days on the network. 
In each time step $t=0, \ldots, T$ we have a budget of $k$ health workers that can visit and actively screen a set of nodes denoted as $\textbf{a}_t \subseteq V$. In real world, there are some people that visit health facilities voluntarily without the active intervention from the health workers. To model such observations, we assume the nodes which turn from $I$ to $S$ state are observed by the health workers at the start of each round. We denote such set as $\textbf{o}_t \subseteq V$. After knowing this information, the agent will select the set of nodes $\mathbf{a}_t$ to screen under the budget constraint $k$. After being screened, the infected patients recover back to the susceptible state, while individuals in the susceptible state remain susceptible. The process repeats until the given time horizon ends. The goal is to maximize the accumulated number of un-infected patients across time, which can be written as $R=\Sigma_{t=0}^{T}\Sigma_{v\in V}  \mathds{1}_{\mathbf{x}^{v}_{t}=S}$ with $\mathds{1}$ being the indicator function. The multi-round active screening is  essentially a sequential decision making problem with combinatorial actions in each time step. %\sj{why do we call this active screening? have we defined this? this seems like a good place to define this. you can say at the beginning that the screening is over T days, in each day $t\in T$ we have a budget of $k$ ... typically k is much smaller than the network. You then can say in passive screening you come up with an assignment of budget ahead of time. Then describe active screening}
%\sj{is there any assumption that we observe the whole state of the network?} \ho{what do you mean? we only know the graph structure and node go from I to S.}

\subsection{Markov Decision Process Formulation}
We start by formulating the active screening problem as a MDP. 
%\sj{we just said it's an MDP. Why are we talking about POMDP? Also please define acronyms. Also the fact that the problem is POMDP/MDP has nothing to do with RL. Remember that this is still the problem formulation not the solution}
%We reduce the problem from a \emph{Partially Observable Markov decision process} (POMDP) to an MDP as POMDP solvers are extremely unscalable with respect to time horizon and state-action spaces, which are not suitable for our case as we have a large time horizon and the state-action space in our problem grows exponentially in screening budget (we assume that the budget grows with respect to the network size). We next describe the components of this MDP.
%\sj{just exponential in budget?}\ho{network size for state as well if we don't replace it with the approximate belief}
%\sj{again this is the formulation -- we should not talk about run time here.} \ho{I wanted to talk about POMDP because we got reviwer mentioning it before, we want to point out that the solver for it is not scalable and thus we formulate the problem as MDP instead for having more scalable solver.} \hp{Seems fine to me.}
%By consider the belief state, defined as the approximate probabilities of each node being infected, as the current state, the problem can be reduced to an MDP formalization as the following:
%\sj{what is a belief state?}

\noindent \textbf{States: } %Graph structure $G$ and our current belief of each individual node state $b_v$.
The hidden state of our problem is the combinatorial health state of each individual node which is partially observable.
To represent the observation uncertainty in the current state, we follow~\citet{ou2020} by defining a \textit{belief state} $b_v$ for every node $v$, which can be interpreted as the approximate probabilities of each node being infected. See Appendix~\ref{app:belief} for the details.
%\sj{are we going to define how we calculate the belief? even if not in the text, I think this should be defined in the appendix for completeness}\ho{I can write that in appendix, it will be very similar to the old paper though}\sj{please add to appendix and mention that we replicate from your old paper for completeness. it's not clear for me that we do a one-step influence propagation or more}

\noindent \textbf{Actions: } Given the current state, the agent can choose any subset of nodes $C \subseteq V$, $|C|\leq k$ to screen. The size of the action space is $\binom{V \setminus \mathbf{o}_t}{k}$ at each round, where  $\mathbf{o}_t$ is the set of nodes that are passively screened (so there is no advantage in actively screening them). %\sj{what is o? what is t? undefined}
%\hp{$\binom{V \setminus \mathbf{o}_t}{k} $ is only a number, not a space. A space should be something like $\{v|v\in xxx \cap |yyy|\leq k \}$}

\noindent \textbf{Rewards: } 
The objective of the active screening problem is to maximize the accumulated number of susceptible nodes. It is natural 
to consider the step wise reward signal as number of susceptible nodes after the active screening, denoted as $r_t=\Sigma_{v\in V}  \mathds{1}_{\mathbf{x}^{v}_{t}=S}$. Since every infected individual has a fixed probability $\gamma$ of being observed by the agent, the step wise reward can be easily estimated.%\sj{so we do not who is infected but we can easily estimate the number of infections given that $\gamma$ percentage of those come to the clinic? we need to emphasize this if this is the case. I thought $\gamma$ percent of the population go to clinic regardless of their health state} \ho{we do mentioned that in the disease model already?}
%\sj{do we need to say something about rewards being partially observable?}

\noindent \textbf{Transitions: } 
In belief of the health states, screened or observed nodes ($v \in \mathbf{o}_t \cup \mathbf{a}_t$) are updated by their ground truth values and the remaining nodes are updated by inferring their posterior probabilities. The key to state transition is the update of belief state, which is defined following~\cite{ou2020}. We refer to Appendix~\ref{app:belief} for the detailed description of belief state update. We also refer to Appendix~\ref{app:notation} for a summary of notations. %The details could be found in appendix~\ref{app:belief}.% \sj{refer to the appendix here}  %\sj{I don't think this is what we need. We want to talk about the transitions of the underlying MDP not our beliefs.}\ho{we use belief as state to reduce POMDP to MDP}

\section{Methodology}
\label{sec:methodology}
Despite a well defined MDP,
it is extremely challenging to solve it due to various reasons. One of the main challenges is that both state space and action space we are facing are high-dimensional. For the state space, even when the true state is available, there are a total of $2^{|V|}$ possible states. When uncertainty is involved, the state values are continuous and therefore the number of states is infinitely large.
%there are infinite continuous states with each state is a probability distribution over the original $2^{|V|}$ possible states. \hp{I removed this distribution thing as we've already decided to treat nodes independently in the state description in Sec. 3}
Furthermore, the states of individual nodes are not independent from each other, but are correlated due to potential contacts from the network. 
For the action space, in each time period we need to choose a combination (subset) of nodes from the entire network.
For a reasonably large network, this combinatorial action space grows exponentially with respect to the screening budget $k$, and becomes intractable as $k$ typically scales as the graph size grows. In the following we show how these challenges are handled in our approach. A summary of notations related to the algorithm is included in Appendix~\ref{app:notation}.

\subsection{Basics of DQN}
Due to the superior performance of RL algorithms in solving large scale MDPs~\cite{mnih2015human,silver2016mastering,silver2017mastering}, we adopt RL as the basis of our solution. More specifically, the backbone of our approach is a hierarchical RL algorithm based on DQN. In this sub-section, we first introduce the basics of RL and DQN, following which we then describe several ideas that further adapt DQN to our formulated problem. We need to emphasize that we do not claim novelty in each of the adaptions, but instead the novelty lies in the innovative way of combining the ideas into solving the particular problem of interest. %We start by introducing the $Q$ value estimation for our high dimensional state action space. We adopt a Deep Q-Network due to its superior performance in solving high dimensional MDPs. %\sj{I am not sure this is theoretically proved anywhere but if you know please add a cite}

%Deep RL has achieved outstanding performance in recent studies. \sj{why do we keep repeating this?}
%\hp{We need to distinguish DQN with RL. DQN is only one type of RL. The description here is only fit for DQN.}
RL is a learning framework where agents learn to perform actions in an environment so as to maximize a certain objective. The two underlying components of RL are the environment, which is defined as the MDP in this paper, and the agent, which represents the learning algorithm. At each time step $t$, the agent takes an \textit{action} based on its \textit{policy} $\pi(\mathbf{a}_t|\mathbf{s}_t)$, where $\mathbf{s}_t$ and $\mathbf{a}_t$ are respectively the state and action of the MDP defined above. The agent then interacts with the environment with the selected action and the environment returns a reward $r_t$ for that action as well as the state $\mathbf{s}_{t+1}$ of the next time step. Q-learning~\cite{watkins1992q} is a value-based RL approach that is based on the notion of Q-function (i.e., state-action value function). The Q-function measures the expectation of accumulated rewards of an action $\mathbf{a}_t$ given state $\mathbf{s}_t$. In the training phase of Q-learning, the policy usually exploits the action with the highest Q-value with a high probability $1-\varepsilon$, and explores random actions with a small probability $\varepsilon$.
The Q-function is typically estimated using the Bellman equation: $Q
^{i+1}(\mathbf{s}_t,\mathbf{a}_t)=r_t+\alpha \max_{\mathbf{a}_{t+1}} Q^{i}(\mathbf{s}_{t+1},\mathbf{a}_{t+1})$, where $i$ indicates the training iteration and $\alpha$ is the discount factor. DQN~\cite{mnih2013playing} improves Q-learning by representing it using deep neural networks, together with other techniques like experience replay over a number of episodes ($\mathcal{E}$), which basically stores the historical training trajectories in a ``replay buffer'' and updates the Q-function by minimizing the loss function $(y-Q(s,a))^{2}$ with batch data from the replay buffer using gradient descent algorithms. Here $y$ is the ``target'' which is estimated using the above Bellman equation (a technique usually called Temporal Difference learning), and $Q(s,a)$ is directly obtained by feeding $s$ and $a$ to the Q-function.

\subsection{GCN-based Function Approximator}
%\hp{Flow: First describe the deficiency of traditional DQN: it does not capture the intrinsic correlations between nodes in a contact network. GCNs are good at this. Therefore we use GCN as the function approximator in Q-function.}\ho{edited}
With the deep neural networks based function approximator, DQN addresses the exponentially large and continuous state space in our formulated MDP. However, one shortcoming of such a function approximator is that it does not capture the intrinsic correlations between node features. Intuitively, the infectious statuses of linked nodes in a social network are inter-dependent. %when dealing with a graph structure state input. 
%The graph information can only be extracted by hand craft and add ass additional feature.

Graph convolutional neural networks (GCNs)~\cite{kipf2016semi} embed the graph structure itself into its network directly, and thus have superior performance on graph type inputs. Each layer of GCN is given by $\mathbf{z}^{l+1}=\sigma (\mathbf{D}^{-\frac{1}{2}}\mathbf{A} \mathbf{D}^{\frac{1}{2}}\mathbf{z}^{l} \mathbf{W})$, in which  $\mathbf{z}^{l}$ is the input of $l$-th layer, $\mathbf{D}$ is the diagonal node-degree matrix that normalizes the adjacency matrix $\mathbf{A}$, $\mathbf{W}$ is the trainable weight matrix and $\sigma(\cdot)$ is the activation function. 
The convolutional layers in GCNs can facilitate the nature of message passing and automatically aggregate the information from neighboring nodes. Such message passing is similar to the infection spread in our epidemic model. The advantage of using GCNs is that we do not need the hand-crafted graph features to represent the information about the graph structure such as the node degrees or eigenvalue of adjacency matrix~\cite{li2012degree}. Inspired by recent advances that combine the power of RL and GCNs-based deep function approximators~\cite{khalil2017learning,qiu2019dynamic,kamarthi2020influence}, we use GCNs in this paper to represent the Q-function.

%Recent advances in the graph embedding techniques have shown great success in solving graph-related combinatorial problems~\cite{khalil2017learning,qiu2019dynamic,kamarthi2020influence} by combining the power of RL and GCN. %The information we have in each state, including graph structure, can be naturally encoded by graph convolutional neural networks (GCNs) \cite{kipf2016semi}.

Our adaptation of the GCNs takes the belief state as input. The action and observation can be naturally encoded in the belief state as we can update the corresponding elements in the belief state vector to their true state. Thus we do not need to encode them as additional features.%\ho{added discription}\hp{now I understand how observation is encoded. how about actions?}\ho{we know both action and observation will become S in next state.} 
We thus combine the observation with the graph structure that is represented as the adjacency matrix $\mathbf{A}$ to our state representation in a very structured way. 
The GCNs learn the underlying graph embedding and automatically form the representation and output the Q-value estimation for our RL agent.%\sj{this is not clear to me. This claim needs a lot of justification. We replaced the state space with something else. We have not yet shown any effectiveness so I am not sure we can have this claim here.}
%Next, we describe how we solve the remaining high-dimensional action space issue.
%\hp{The formulas about GCN are missing. You may write something very similar to Eq.3 in my IJCAI paper~\cite{qiu2019dynamic}.} \ho{You mean the GCN layer equation $y^{l+1}=\sigma (D^{-\frac{1}{2}}A D^{\frac{1}{2}}y^{l} \theta)$? We can briefly describe these symbol.}\hp{yes. By writing this down it is clear what a GCN layer is.}

\subsection{Sequence of Sequence Framework}
In addition to the challenges that arise from the high-dimensional and graph-structured state space, as described previously, another challenge is the combinatorial action space in each time step of screening. 
To address this challenge, we propose a hierarchical RL approach by re-formulating each time step in the original MDP as a sub-sequence of decisions by itself.
We call this framework sequence of sequence (SOS). We refer to the original multi-rounds of active screening as \textit{time} sequence. Correspondingly, we refer to the sub-sequence problem of selecting $k$ nodes 
in each round as the \textit{budget} sequence. In each of the budget sequence, we are solving a separate sequential decision making problem with a final reward $R_t$, a finite time horizon of $k$, and an action space $V \setminus \mathbf{o}_t$ whose size is equal to the network size.
%The idea behind SOS is that each budget sequence can be viewed as a separate sequential decision-making process with a final reward $R_t$ and a finite time horizon of $k$ with action space $V \setminus \mathbf{o}_t$ whose size is equal to the graph size. 

The SOS framework allows for a tractable action space which can be used by an RL algorithm. 
However, two additional issues arise in such conversion of the action space. First, although the states and actions are well represented by GCNs, the algorithm does not take into account the remaining budget at the budget sequence. In fact, the policies should be very different when there is plenty of budget left versus little budget left. Intuitively, this is because the actions taken when there is more budget left should consider more about its future effect, while actions taken when there is less budget left tend to be more myopic. Therefore, a single-agent framework in the budget sequence, which treats all states equally for different remaining budget, usually does not work well.
%\hp{how about the information about time steps in the time sequence? is it important? if we did not take that into account, we need to be prepared to answer in the rebuttal.} \ho{If you mean the remaining time I don't think it is very important.}
%\hp{I don't get the main point. Are you trying to say we need to design a different secondary agent for each sub-time-step in the budget sequence? }\ho{This is trying to clarify that the secondary agents are not myopic that only care within the time series to address the possible confusion shahin raised earlier.}
%\hp{I think the point here should be to highlight the issue that remaining budget information should be taken into account. I re-wrote this part. please take a look if it is accurate.}\ho{looks convincing}
Second, by introducing the SOS framework, only the final step in the $k$ budget-steps for the budget sequence gets a reward signal. The sparseness of reward is known to slow the convergence of RL~\cite{rlblogpost}. Therefore distributing the reward $R_t$ to each action in the budget sequence for proper reward signaling is a non-trivial task. %This is important since effective reward signals help faster convergence for RL algorithms. 
%\sj{I do not understand this challenge. Can we just feed the remaining budget as a part of state space to the budget sequence? in any way this paragraph needs major clarification} \ho{Show what if no secondary sequence happens.} \ho{Combine 4.3 and 4.4??}

%\sj{not convincing. when we solve for the budget sequence with RL, we only consider future actions in that sequence and not the time sequence. so in some sense we are myopic. I am not convinced with this description} \ho{Explain it is not myopic in content}

\begin{figure*}[h]
\centering
\includegraphics[width=0.6\textwidth,keepaspectratio]{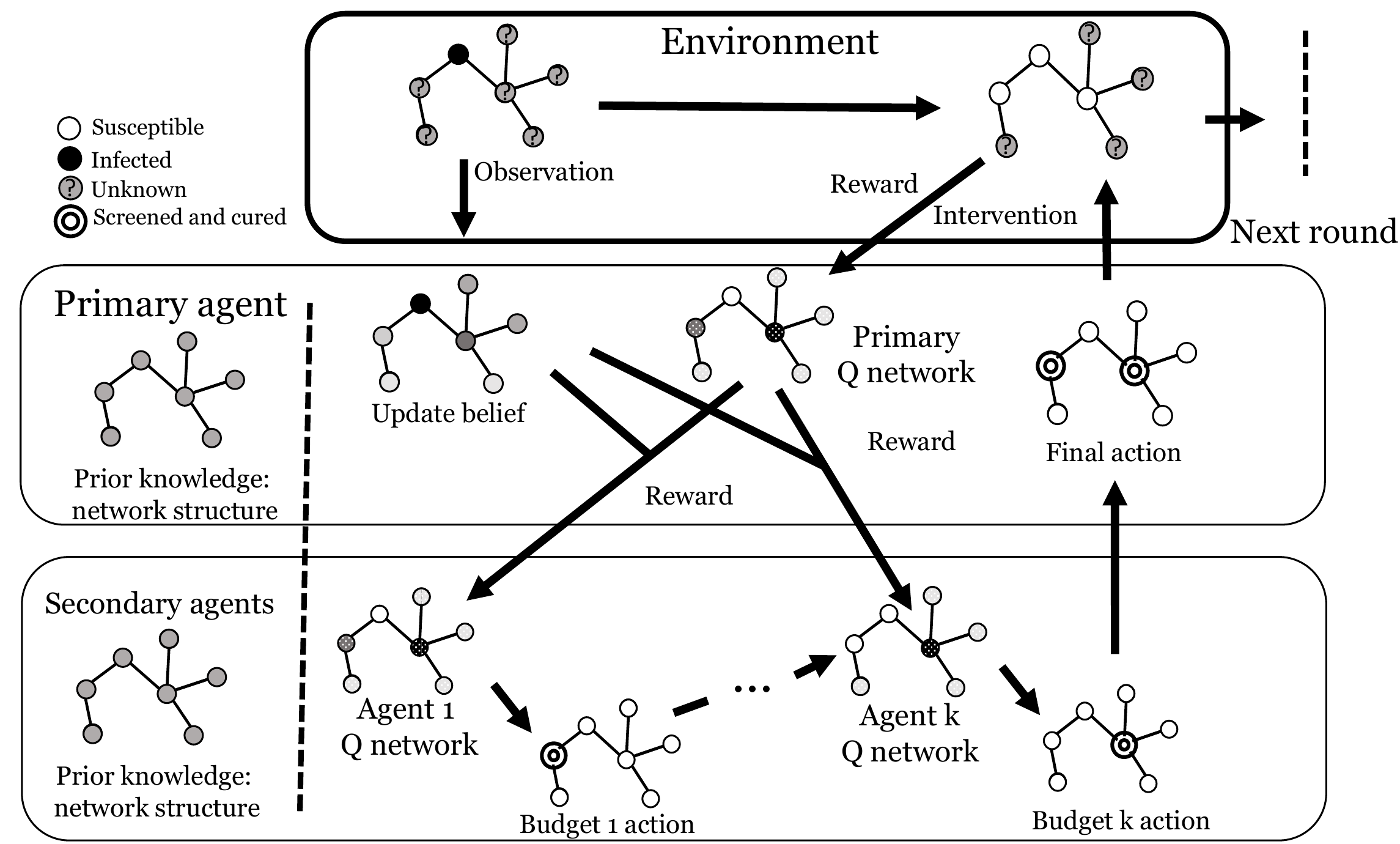}%
\caption{The overall flow of one round decision making in a budget sequence using our two-level RL structure. The top row depicts the environment, the second row is the workflow of the primary agent, and the third row is the workflow of the $k$ secondary agents.
The primary agent starts by observing the initial state of the environment. It then updates the belief of the state. The belief state is passed down to the first secondary agent, which then evaluates the value of each feasible node action using its own Q network and decides which node to select. The successive secondary agents work sequentially until the $k$-th secondary agent, i.e., when budget $k$ is spent. The selected actions are collected to get the final action set $\mathbf{a}_t^I$ which is uploaded to the primary agent. The primary agent then interacts with the environment using this action and gets reward signals.
%The primary agent manages the belief update, delivers action and receives reward from the environment, and maintains the long term reward estimation of the time sequence. The secondary agents receive reward from the primary agent. Each of them is responsible for choosing one node in the budget sequence. %\sj{do we ever say what colors, double circles and question marks refer to? the figure is small especially the text}\ho{add the illustration}\ho{add back to the text}
}\label{RL_plot}
\end{figure*}

\subsection{Primary and Secondary Agents}
Inspired by hierarchical RL~\cite{dayan1993feudal,parr1998reinforcement,sutton1999between,dietterich2000hierarchical}, we propose a multi-agent RL approach with a two-level structure that manages the time and budget sequences in a hierarchical way. The overall flow of the two-level structure is depicted in Fig.~\ref{RL_plot}. The idea is to capture the remaining budget information by having $k$ secondary agents, where each secondary agent maintains a policy for a different remaining budget value.\footnote{An alternative is to train a single agent and encode the remaining budget information directly as part of the state. However, we show via ablation study in Appendix~\ref{app:ablation} that this leads to sub-optimal solution quality.} The primary agent, as shown in Algorithm~\ref{alg:MS}, manages the time period reward signal in a given time step of the time sequence. The reward signal acts as the secondary agents' total reward in the budget sequence and is distributed over the $k$ secondary agents. In the following, we use superscripts $I$ and $II$ to distinguish concepts that correspond to the primary and secondary agents.

\noindent \textbf{Primary Agent}
In the time sequence, the primary agent works as follows. First, it receives the combinatorial action and the secondary agents' memories that store the trajectories from the secondary agents (line~\ref{al1}). Passing this action to the environment, it receives the observation and reward (line~\ref{al2}). It then handles the new state representation of the next time step (lines~\ref{al3} and~\ref{al4}). Finally, it updates both primary and secondary agents' memories ($\mathcal{M}^I$ and $\mathcal{M}^{II}$) (lines~\ref{al5} and~\ref{al6}).  Each element of $\mathcal{M}^I$ and $\mathcal{M}^{II}$ is basically a tuple of state, action, next state and rewards that will be used to train the primary and secondary agents' Q-functions. Note that the state of the secondary agents is slightly different from that of the primary agent, which will be explained in the next paragraph.
These memories are used to update the GCN-based Q-functions by minimizing the loss function $(y-Q(s,a))^{2}$ through gradient descent (lines~\ref{alg_fit1} and~\ref{alg_fit2}), where $y$ is the target. At each time step $t$, the target of each agent is given by:
\begin{align}\label{Q_fit}
y^I=& r^{I}_{t} + \alpha Q^I(\mathbf{s}_{t+1}^I, \mathbf{a}_{t+1}^I), \\
y^{II}_{i}=&r^{II}_{i}+\alpha Q^{II}_{i+1}(\mathbf{s}^{II}_{i}, a_{i+1}^{II})  \text{ for } i=1, \ldots,  k-1,\\
y^{II}_{k}=& r^{II}_{k} + \alpha Q^{I}(\mathbf{s}_{t+1}^{I}, a_{t+1}^{I}). \label{Q_fit_e}
\end{align}
Note that in lines~\ref{ln:cur} and~\ref{ln:cur2}, the belief state and reward are obtained using the idea of curriculum learning that mitigates the reward sparseness issue for the secondary agents. We will elaborate this idea in the next subsection.

\begin{algorithm}[ht]
\caption{\textsc{Primary Agent}}\label{alg:MS}
% \begin{flushleft}
% \textbf{Input}: $\mathbf{A},T, \mathcal{E},\mathcal{M}^M,\mathcal{M}^W)$\\
% \textbf{Output}: $Q^M$
% \end{flushleft}
\begin{algorithmic}[1] %[1] enables line numbers
\FOR{$episodes = 1,...,\mathcal{E}$} 
\STATE Initialize and acquire initial belief ($\mathbf{b}_0$) and observation ($\mathbf{o}_0$)
\STATE $\mathbf{s}_0^I \gets \textit{Graph Embeding}(  \mathbf{A}, \mathbf{b}_0)$
\FOR{$t \in 0,...,T$} 
\STATE $\tilde{\mathbf{b}}_t \gets \textit{Curriculum Belief Transform} (\tau, \mathbf{b}_t$) \label{ln:cur}
\STATE  $\mathbf{a}_{t}^I, m^{II} \gets \textit{Secondary Agent} (\tilde{\mathbf{b}}_t,Q^I)$\label{al1}
\STATE $\mathbf{o}_{t+1}, r^{I}_{t} \gets \textit{Environment}(\mathbf{a}_{t}^I) $\label{al2}
\STATE $\tilde{r}^{I}_t \gets \textit{Curriculum Reward Transform} (\tau, r^{I}_t)$ \label{ln:cur2}
\STATE $\mathbf{b}_{t+1} \gets \textit{Belief Update}(\mathbf{b}_{t},\mathbf{o}_{t},\mathbf{a}_{t}^I) $\label{al3}
\STATE $\mathbf{s}_{t+1}^I \gets \textit{Graph Embeding}(  \mathbf{A}, \mathbf{b}_{t+1})$\label{al4}
\STATE $\mathcal{M}^I \gets \mathcal{M}^I \cup \left\{(\mathbf{s}_{t}^I,\mathbf{a}_{t}^I,\tilde{r}^{I}_t,\mathbf{s}_{t+1}^I)\right\}$\label{al5}
\STATE $\mathcal{M}^{II} \gets \mathcal{M}^{II} \cup m^{II}$\label{al6}
\ENDFOR
\STATE Decrease $\tau$
\ENDFOR 

\STATE $\textit{Fit } Q^I \textit{with regressor net using } \mathcal{M}^I$ \label{alg_fit1}
\STATE $\textit{Fit } Q^{II}_{0} ... Q^{II}_{k-1} \textit{with regressor nets using corresponding } \mathcal{M}^{II}$ \label{alg_fit2}
\end{algorithmic}
\end{algorithm} 

\noindent \textbf{Secondary Agents} As for the budget sequence, instead of training a single agent and performing a batch selection of nodes, we train $k$ secondary agents to handle each budget sequentially. As described above, the purpose of doing so is to differentiate secondary agents who know there is plenty of budget left and those who know there is little budget left, so they could learn different policies.
The state $\mathbf{s}_i^{II}$ of a secondary agent $i$ is obtained by encoding the primary agent's action (i.e., the set of nodes selected so far) taken upon the state of the previous budget step. The action $a^{II}_i$ for each secondary agent $i$ is to choose one node to add to the primary agent's action set $\mathbf{a}^{I}$.
$\mathbf{a}^{I}$ is initialized as an empty set (in line~\ref{ln:secondary_start}) and will be updated by appending actions from each secondary agent.
%\sj{Why cannot we have the remaining budget as a part of state space?} 
%\hp{this should be mentioned at the beginning of sec. 4.1, so people know there are multiple secondary agents}\ho{added in the beginning of sec 4.4}\hp{there is no definition of state and action etc. for the secondary agent. readers who don't know what we are doing will get lost} \ho{edit by adding the symbol we used, hope that it will make it clearer}
In the for loop that represents a budget sequence (from line~\ref{ln:for_start} to line~\ref{ln:for_end}), each secondary agent $i$ will select the action $a_i^{II}$ that maximizes its Q-function $Q^{II}_i(s_i^{II},a_i^{II})$ and add it to the primary agent's action set $\mathbf{a}^{I}$ (lines~\ref{ln:act_select} and~\ref{ln:act_append}). 
%Each network is responsible for different part of the action choice. 
%as shown in Line 6 of Alg. 2, the reward signals of the secondary agents 
After that, it receives the reward in line~\ref{ln:secondary_reward}, which is obtained from the primary agent (as a proxy) in a temporal difference learning manner. 
Next, the secondary agent will encode the primary agent's action $\mathbf{a}^{I}$ (i.e., the set of nodes selected so far) into its current state $\mathbf{s}_i^{II}$, which is used as next state $\mathbf{s}_{i+1}^{II}$ and pass this information to the next secondary agent (line~\ref{ln:secondary_state_update}). 
Finally, it stores the above information as memory, so it can be used later to update the Q-functions in Equations~\eqref{Q_fit}-~\eqref{Q_fit_e}. For extremely large graphs, we reduce the memory and computation time by assigning fewer than $k$ secondary agents, where each secondary agent is responsible for a portion of the budget instead. For example, for a budget of 20, if we assign 10 secondary agents, each secondary agent needs to select $20/10=2$ nodes at a time.

\begin{algorithm}[h]
\caption{\textsc{Secondary Agents} }
\begin{flushleft}
\end{flushleft}
\begin{algorithmic}[1] %[1] enables line numbers
\STATE $\mathbf{a}^{I}, m^{II}_{1}...m^{II}_{k} \gets \emptyset$ \label{ln:secondary_start}
\STATE $\mathbf{s}^{II}_{0}\gets \textit {Encoding}(\mathbf{s}^{I},\mathbf{a}^{I}) $
\FOR{$i \in 1,...,k$} \label{ln:for_start}
\STATE $a_{i}^{II} \gets \arg\max Q^{II}_{i}(s^{II}_{i},a_{i}^{II})$\label{ln:act_select}
\STATE $\mathbf{a}^{I}\gets \mathbf{a}^{I} \cup a_{i}^{II}$\label{ln:act_append}
\STATE $\mathbf{r}^{II}_{i} \gets Q^{I}(\mathbf{s}^I, \mathbf{a}^{I})-Q^{I}(\mathbf{s}^I, \mathbf{a}^I\setminus a_i^{II})$\label{ln:secondary_reward}
\STATE $\mathbf{s}^{II}_{i+1} \gets \textit {Encoding}(\mathbf{s}^{I},\mathbf{a}^{I}) $\label{ln:secondary_state_update}
\STATE $m^{II}_{i} \gets m^{II}_{i} \cup \left\{(\mathbf{s}^{II}_{i},a_{i}^{II},\mathbf{r}^{II}_{i},\mathbf{s}^{II}_{i+1})\right\}$\label{ln:for_end}
\ENDFOR 
\RETURN $\mathbf{a}^I, m^{II}$
\end{algorithmic}
\end{algorithm}

\subsection{Curriculum Learning}
As discussed earlier, a major issue with the two-level framework is the sparseness of rewards for the secondary agents. Inspired by curriculum learning~\cite{bengio2009curriculum}, we address this by incrementally increasing the complexity of the learning tasks for the secondary agents. 
This is called \textit{Curriculum Transformation} in lines~\ref{ln:cur} and~\ref{ln:cur2} in Algorithm~\ref{alg:MS} and is described as the following equations:
\begin{align}
\tilde{\mathbf{b}}_t=& \tau \mathbf{x}_t +(1-\tau)\mathbf{b}_t,  \label{eq:CT_state}\\
\tilde{r}^{I}_t=& \tau \bar{r}_t +(1-\tau)r^{I}_t,\label{eq:CT_reward}
\end{align}
where $\tilde{\mathbf{b}}_t$ and $\tilde{\mathbf{r}}_t^I$ are respectively the belief state and reward of the primary agent (used to update the target values for both the primary and secondary agents) after curriculum transformation. $\tau$ is an auxiliary coefficient that gradually decreases from 1 to 0 in the first few epochs of training. It adjusts task difficulties from a relatively easier problem ($\tau=1$) to the original problem ($\tau=0$). 
At the early stage of training ($\tau=1$), we warm up the learning by feeding the algorithm with the true state information $\mathbf{x}_t$ (Eq.~\eqref{eq:CT_state}). Moreover (Eq.~\eqref{eq:CT_reward}), we set the reward to be $\bar{r}_t=\sum\nolimits_{v\in \mathbf{a}_t^I}  \mathds{1}_{b^{v}_{t}=S}$, which means the total number of infected nodes in the action set $\mathbf{a}_t^I$, instead of in the total number of susceptible nodes $S$. In this way, the algorithm learns to greedily cure nodes that are infected. As the training continues, decreasing the auxiliary coefficient $\tau$ takes two effects. First, it gradually removes the true state information. It is worth noting that the true state is only used in warming up the training, and is not used during testing. 
Second, it shifts the reward from being constrained in the set of infected nodes to the true reward, and explores potentially more optimal actions outside the set of infected nodes.
When $\tau$ is 0, the belief and reward become identical to the original problem ($\tilde{\mathbf{b}}_t=\mathbf{b}_t$  and $\tilde{r}^{I}_t=r^{I}_t$).%\sj{why $\tau = 1$ is easy? what is $\bar{s}$ and $\bar{r}$?}\ho{fixed}
%We introduce an auxiliary coefficient denote as $\tau$, which gradually decrease from $1$ to $0$ in the first few epoch of training. It adjusts tasks' difficulties from a relatively easier task ($\tau=1$) to the original problem ($\tau=0$). \sj{I am not following this idea. Is this similar to decreasing learning rate? where in the training this $\tau$ parameter is used?}\ho{Now the high level is discussed first, the $\tau$ is explained in the next para.}

%At the early stage of training, we warm up the learning by setting the reward to success rate, which means the total number of infected nodes in the action set ($v$ for $v\in\mathbf{a}_t$ and $\mathbf{s}^{v}_{t}=I$), instead of the total number of susceptible nodes ($v$ for $v\in V$ and $\mathbf{s}^{v}_{t}=S$). 
%\sj{why does this help? do we need to use the term hit rate?}\ho{we make sure the algorithm knows it should cure I nodes in the start of training.}
%Also, we allow the algorithm to use the true state $\mathbf{x}_t$. 

\section{Experiments}
\label{sec:exp}

\textbf{Datasets} We evaluate the effectiveness of our proposed approach on different real-world contact networks. These datasets are open-sourced and are also used in ~\citet{ou2020} which is the closest related work to ours. The nodes in each of these networks represent human individuals, the edges represent different forms of contact. We next provide a brief description for each of the datasets. (i) \textbf{Hospital}~\cite{hospital}: A contact network collected by wearable devices in a university hospital. A link is built if close range interactions with long enough duration is detected.
(ii) \textbf{India}~\cite{jackson}: A network collected from one of the villages in India by households surveying. The edges represent the real world social contact.
(iii) \textbf{Face-to-face}~\cite{infectious}: A network describing face-to-face behavior during the exhibition ``INFECTIOUS: STAY AWAY'' in 2009 at the Science Gallery in Dublin. The network simulates the close contact of individuals that influenza might spread through.
(iv) \textbf{Flu}~\cite{stanfordflu}: A network of close proximity interactions in an American high school sampled in an ordinary day. The network is collected using wireless sensor network technology. An edge is built for close physical proximity.
(v) \textbf{Irvine}~\cite{panzarasa2009patterns}: A network collected from an online student community in UC Irvine to analyze spread of information or rumour using epidemic models. The edges represent the communication over online messages.
% \begin{itemize}
% \item[(1)] \textbf{Hospital}~\cite{hospital}: A contact network collected by wearable devices in a university hospital. A link is built if close range interactions with long enough duration is detected.
% \item[(2)] \textbf{India}~\cite{jackson}: A network collected from one of the villages in India by households surveying. The edges represent the real world social contact.
% \item[(3)] \textbf{Face-to-face}~\cite{infectious}: A network describing face-to-face behavior during the exhibition ``INFECTIOUS: STAY AWAY'' in 2009 at the Science Gallery in Dublin. The network simulates the close contact of individuals that influenza might spread through.
% \item[(4)] \textbf{Flu}~\cite{stanfordflu}: A network of close proximity interactions in an American high school sampled in an ordinary day. The network is collected using wireless sensor network technology. An edge is built for close physical proximity.
% \item[(5)] \textbf{Irvine}~\cite{panzarasa2009patterns}: A network collected from an online student community in UC Irvine to analyze spread of information or rumour using epidemic models. The edges represent the communication over online messages.
% \end{itemize}
These networks display diversity as captured by parameters such as network size $|V|$, spectral radius $1/\lambda_A^*$, average degree $d$,  average shortest path length $\rho_L$ and assortativity $\rho_D$ as depicted in Table~\ref{resultstable}. 
\begin{table}[ht!]
\caption{Properties of the contact network datasets. }
\scalebox{1}{\begin{tabular}{|l|ccccc|}
\hline
Network  & $|V|$ & $\frac{1}{\lambda_A^*}$ & $d$ & $\rho_L$ & $\rho_D$\\
\midrule\midrule
\textbf{Hospital}~\cite{hospital} & 75 & 0.027 & 30.37 & 1.60 &-0.18 \\
\textbf{India}~\cite{jackson} & 202 & 0.095 & 6.85 & 3.11 & 0.02   \\
\textbf{Face-to-face}~\cite{infectious} & 410 & 0.042 & 13.49 & 3.63 & 0.23 \\
\textbf{Flu}~\cite{stanfordflu} & 788 & 0.003 & 300.23 & 1.62 & 0.05 \\
\textbf{Irvine}~\cite{panzarasa2009patterns} & 1899 & 0.021 & 14.57 & 3.06 & -0.18 \\
\hline
\end{tabular}}
\label{settingstable}
\end{table}

\noindent \textbf{Experimental setting} In all experiments, we fix the passive screening rate $\gamma$ to $0.05$. 
Due the the high diversity of the networks, it is difficult to find one fixed transmission rate $\beta$ that is suitable for every network. A fixed transmission rate is either too high for the dense networks so that the whole network will become infected no matter what policy is deployed, or too low for the sparse networks so that the disease will be eradicated without any intervention. 
We thus adjust the transmission rate according to the density of the network. Specifically, we adjust $\beta$ based on the spectral radius $1/\lambda_A^*$, also known as the epidemic threshold, where $\lambda_A^*$ is the largest eigenvalue of the network adjacency matrix.
%\sj{what does the epidemic threshold mean?}
If there are no additional interventions, the disease will not be eradicated eventually if and only if $ \beta > \gamma/ \lambda^{*}_A$~\cite{wang2003epidemic}. %\sj{cite please. also rewrote the condition to make it look simpler. please make it is correct}
%\sj{are you talking about interventions?}
We set $\beta= 10 \gamma/ \lambda^{*}_A$, which is $10$ times the value corresponding to the epidemic threshold. % \sj{should the last two $A$ be $A^*$?}
%\hp{why 10 times?}\ho{I don't have a good reason for this 10 times and 0.1 for budget.} 
We further set $k=0.1|V|$ for the active screening budget in each time period. Finally, we set the total time horizon to $T=100$. All the results presented are averaged over $30$ trials. %\sj{Is this large enough in the epi sense? how can we justify} %to show our algorithm is able to apply to scenario where large time horizon is required like rapidly spread disease.
%\hp{how about the parameters for the algorithms?}\ho{you mean learning rate etc in RL?}\hp{yes that's about the training, and about neural networks architectures like how many layers of GCN we use, what are the dimension of the tensors in each layer.}
For our RL approach, we set the future discount factor $\alpha$ to $0.98$, exploration probability in the epsilon greedy approach is $0.1$ and the learning rate is $0.005$. We trained the RL agents for $100$ episodes for $100$ iterations of refits. The memory capacity of the relay buffer is set to $5000$ tuples for each agent. We used the sigmoid activation function. There are four layers of sizes $8$,$16$,$8$ and $32$. For all graphs except the largest one, the training finishes within 3 hours on a laptop with 6 cores, 2.60 GHz intel CPU, and 16 GB RAM. For the largest graph, \textit{Irvine} network, it takes about one day to finish on the same laptop and is significantly shortened after using an HPC.\footnote{Codes and Data can be found in https://bit.ly/32wtsEk}

\noindent \textbf{Baselines} We simply call our approach RL. The baselines we are testing against are (i) \textit{Eigenvalue} (greedily choosing nodes that decrease the largest eigenvalue of the remaining sub-graph after removal until the budget is exhausted), (ii) \textit{MaxDegree} (choosing $k$ nodes with the largest degrees), (iii) \textit{Random} (randomly selecting nodes), (iv) \textit{Full-REMEDY} (the algorithm in~\cite{ou2020} that is un-scalable to large time horizons)  and (v) \textit{Fast-REMEDY} (the scalable version of \textit{Full-REMEDY} that does not account for the future effect of actions).

% \begin{table}
% \centering
% \caption{Average improvement over no intervention for $30$ testing iterations. Note that as will be shown in the next subsection, \textit{Full-REMEDY} does not scale to $T=100$ and thus its results are not included. The numbers in the brackets are improvements over the best alternative \textit{Fast-REMEDY}.}
% \scalebox{0.7}{

%  \begin{tabular}{|l|c c c c c|} 
%  \toprule
%  \multirow{2}{*}{Network} &  \multicolumn{5}{|c|}{Reduction in infections compared with no intervention} \\ \cmidrule{2-6}
%  &Eigenvalue & Max-Degree&Random&Fast-REMEDY & \textbf{RL}\\
%  \midrule\midrule
%  \textbf{Hospital} & 1019 & 1015 & 2344 & 3837  & \textbf{4196(9.4$\%$)} \\
%  \textbf{India} &  2668 & 3115 & 6388 & 10033  & \textbf{11270(12.3$\%$)} \\ 
%  \textbf{Face-to-face} & 4225 & 4705 & 8948 & 9919 & \textbf{13283(33.9$\%$)} \\
%  \textbf{Flu} &  7706 & 7725 & 9636  & 10298 & \textbf{11743(14.0$\%$)}\\
%  \textbf{Irvine} & 48490 & 49163  & 42277 & 53159 & \textbf{65128(22.5$\%$)} \\
%  \bottomrule
%  \end{tabular}}
% \label{resultstable}
% \end{table}

\begin{table}
\centering
\caption{Average improvement of different algorithms. \textit{Full-REMEDY} does not scale to $T=100$. Thus its results are not included. The numbers in the brackets are improvements over the best alternative \textit{Fast-REMEDY}.}
\setlength{\tabcolsep}{3.8pt}
\scalebox{0.7}{
 \begin{tabular}{|l|c c c c c|} 
 \toprule
 \multirow{2}{*}{Network} &  \multicolumn{5}{|c|}{Reduction in infections compared with no intervention} \\ \cmidrule{2-6}
 &Eigenvalue & Max-Degree&Random&Fast-REMEDY & \textbf{RL}\\
 \midrule\midrule
 \textbf{Hospital} & 1019$\pm$99 & 1015$\pm$131 & 2344$\pm$136 & 3837$\pm$340  & \textbf{4196$\pm$416 (9.4\%)} \\
 \textbf{India} &  2668$\pm$258 & 3115$\pm$292 & 6388$\pm$259 & 10033$\pm$518  & \textbf{11270$\pm$501 (12.3\%)} \\ 
 \textbf{Face-to-face} & 4225$\pm$211 & 4705$\pm$219 & 8948$\pm$173 & 9919$\pm$499 & \textbf{13283$\pm$301 (33.9\%)} \\
 \textbf{Flu} &  7706$\pm$190 & 7725$\pm$203 & 9636$\pm$163  & 10298$\pm$460 & \textbf{11743$\pm$503 (14.0\%)}\\
 \textbf{Irvine} & 48490$\pm$298 & 49163$\pm$378  & 42277$\pm$270 & 53159$\pm$673 & \textbf{65128$\pm$781 (22.5\%)} \\
 \bottomrule
 \end{tabular}}
\label{resultstable}
\end{table}

\begin{figure}[ht!]
\includegraphics[width=0.4\textwidth,keepaspectratio]{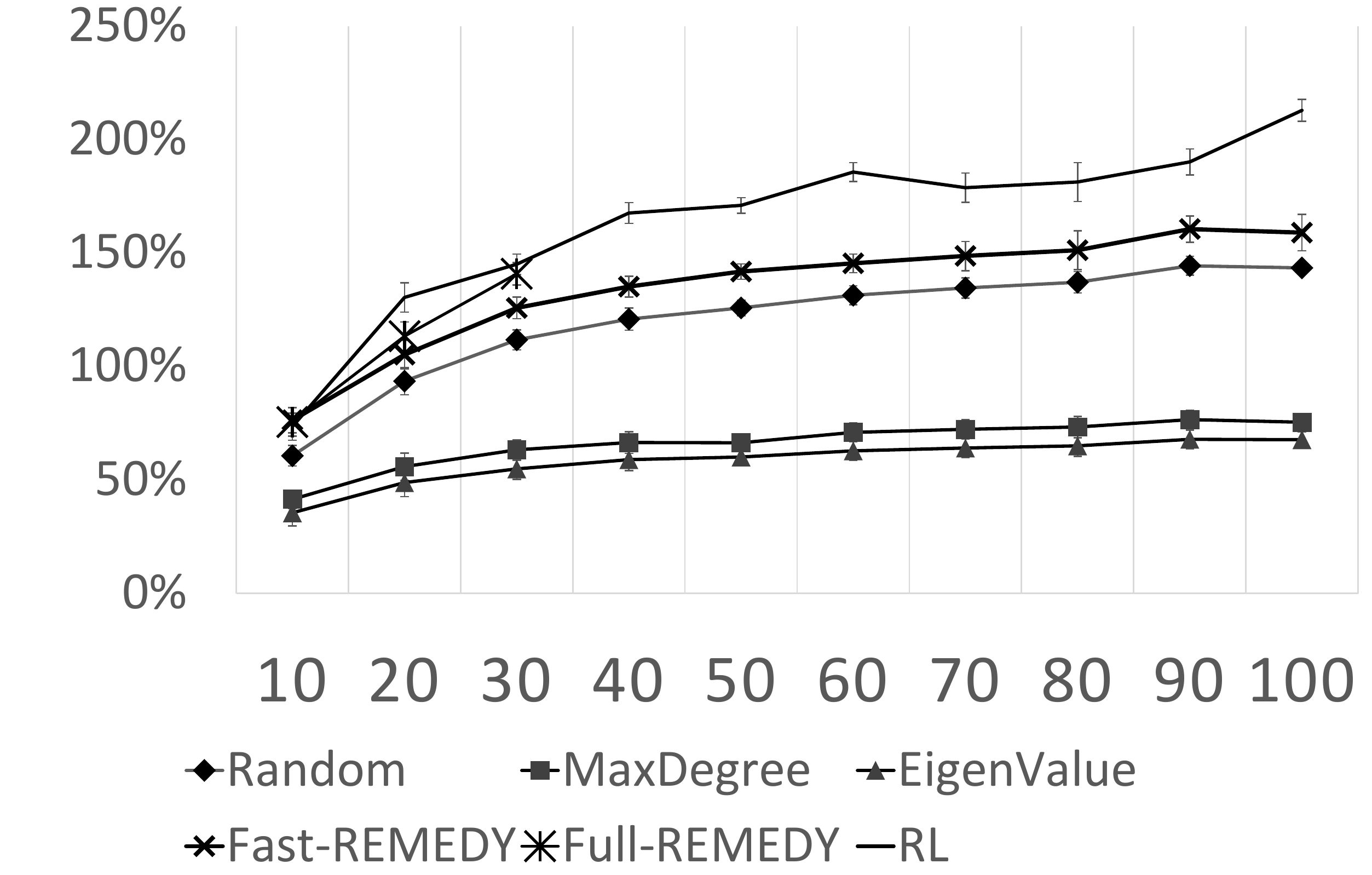}%
\centering
\caption{The performance of each algorithm for different time horizons in the Face-to-face network. The x-axis is the time horizon and the y-axis is the improvement of solution quality over no intervention.
}\label{time_plot}
\end{figure}

\subsection{Solution Quality} Table~\ref{resultstable} shows the increase in average reward compared with no intervention. 
We can see that our approach outperforms the state-of-art (i.e., \textit{Fast-REMEDY}) by a margin of $9- 33\%$. 
There are a few observations worth noting. First, we are implementing the same baselines on the same datasets with significantly larger time horizon compared with~\citet{ou2020}. The ranking of these baselines is consistent with~\citet{ou2020} that are run over a shorter time horizon of 10. Furthermore, results on \textit{Hospital} and \textit{Flu} show a more significant difference as we adjust the transmission rate according to the graph density. Second, \textit{Eigenvalue} and \textit{MaxDegree} baselines perform similarly to each other when we increase the time horizon where as in~\citet{ou2020}, \textit{MaxDegree} clearly outperforms \textit{Eigenvalue} when the time horizon is short. This is expected as the \textit{Eigenvalue} baseline is aimed for long term disease eradication by increasing the epidemic threshold and thus preforms better in the long term. Finally, in \textit{Face-to-face} and \textit{Irvine} networks, our approach performs significantly better compared with the best baseline \textit{Fast-REMEDY}. Interestingly, these are also the networks where \textit{Full-REMEDY} outperforms \textit{Fast-REMEDY} in~\citet{ou2020}. In these networks, the algorithms can benefit more by looking ahead compared with other networks. %\sj{again, this needs some explanation. We cannot hypothesize here. What are the differences between the networks? Maybe node selection analysis can help us gain insight} %\sj{how do you know full-remedy outperformed fast-remedy and if we know that why full-remedy is not included?} \ho{good point, it out perform it in the original paper. We might show it on that table with different T I showed you guys on smaller T for one graph.}

%\sj{This section needs some work. do we have any insight on what policy is used by fast-remedy and how does the policy change when using RL?} \ho{I'm also thinking if there's way to visualize or analyze this.}

\subsection{Scalability Against Time Horizon}

%\sj{did we motivate why having a longer time horizon is valuable in practice for any of the disease that we motivate with in the intro?}
To study how well our approach scales against planning time horizon compared with baselines, we conduct experiments on various time horizons ranging from $10 - 100$. Figure~\ref{time_plot} shows the performance on the face-to-face network for each algorithm on the y-axis when varying the time horizon on the x-axis. \textit{Full-REMEDY} does not scale over time horizons longer than $30$ even on a high performance computer and our approach is in par or better with its performance in these short time horizons. We pick face-to-face network as an example and similar trends can be observed for all the networks in Appendix~\ref{app:exp}. Particularly, in the largest network Irvine, our approach scales 10 times of that in \textit{Full-REMEDY}, meanwhile with better solution quality compared with \textit{Fast-REMEDY}.

\begin{figure}[ht!]
\includegraphics[width=0.4\textwidth,keepaspectratio]{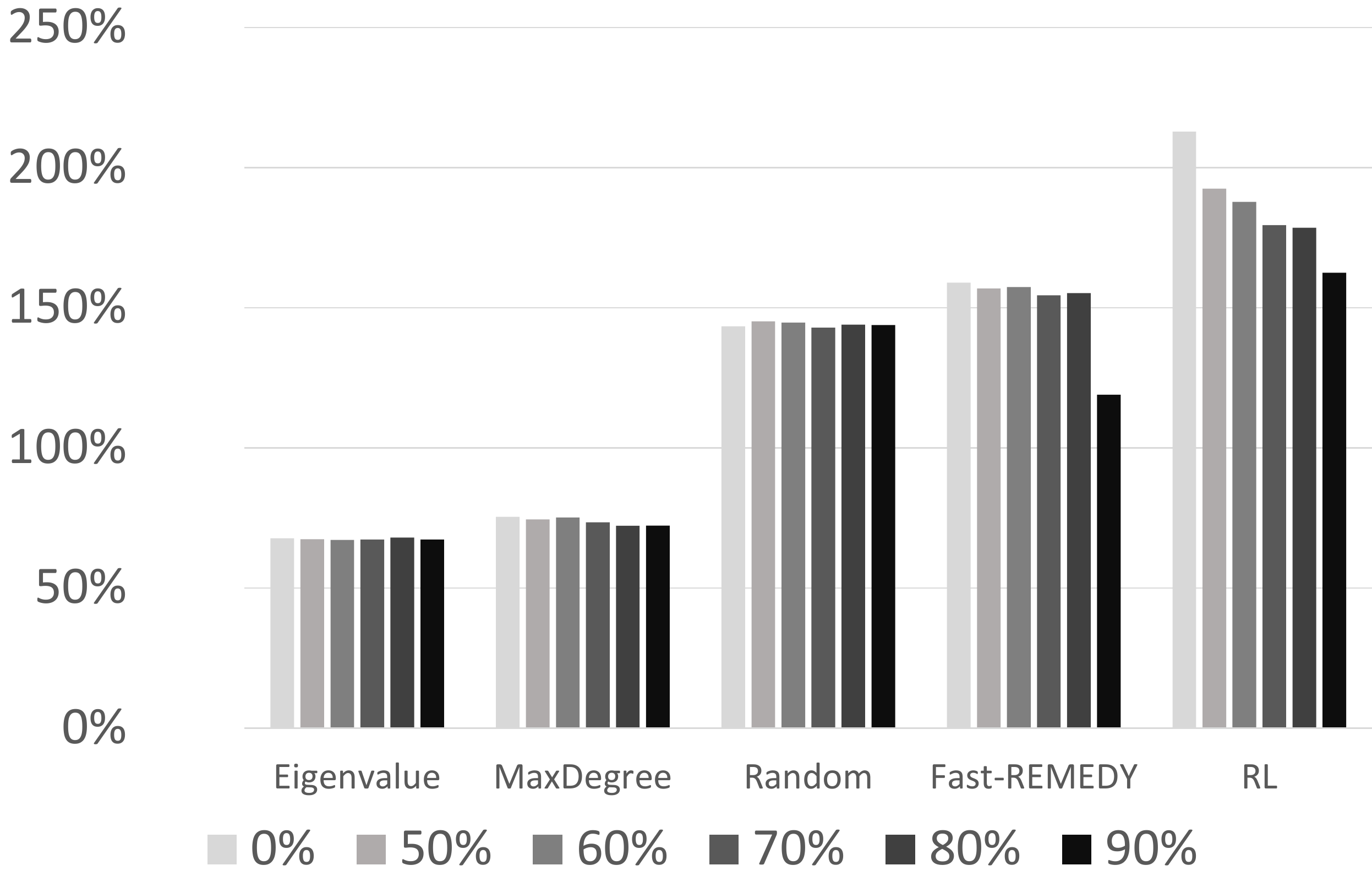}%
\centering
\caption{The performance of each baseline for different edge removal fractions. The x-axis indicates the improvement over no intervention. %and y-axis indicates the percentage of removed edges.
}\label{uncertain_edge}
\end{figure}

\begin{figure}[ht!]
\includegraphics[width=0.4\textwidth,keepaspectratio]{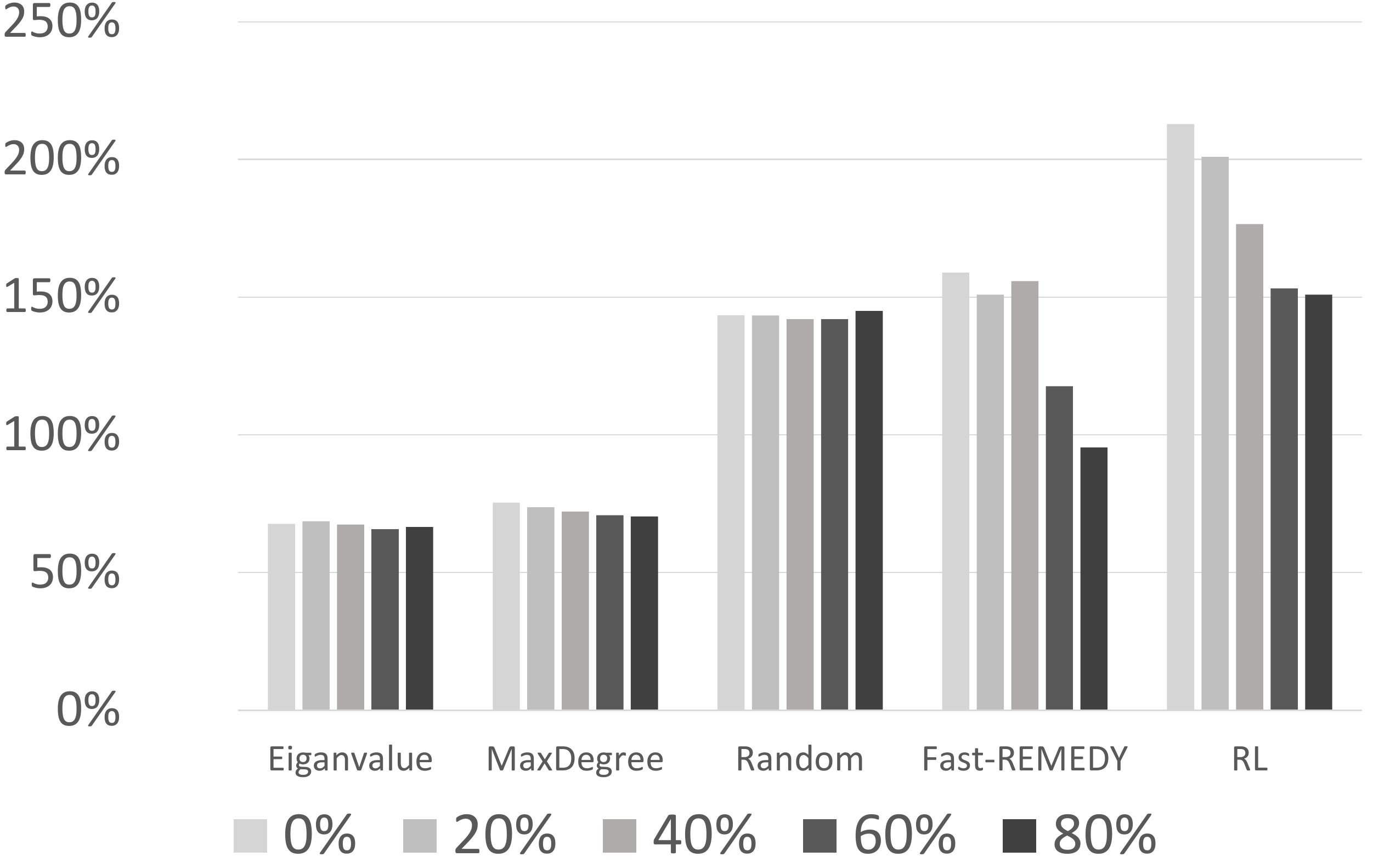}
\centering
\caption{The performance of each baseline for different node removal fractions. The x-axis indicates the improvement over no intervention. %and y-axis indicates the percentage of nodes whose adjacent edges are removed.
}\label{uncertain_node}
\end{figure}

\subsection{Robustness Against Structure Uncertainty}
Although we assume perfect knowledge of graph structure (sans the infectious state of the individuals), one of the main obstacles in implementing active screening in practice is the lack of this perfect knowledge. To evaluate how robust different methods are against structural properties, we design and run the methods on two models for structural uncertainty. In the first one, we assume a constant fraction of the edges are unobserved where this fraction is a parameter. In the second one, we assume all the edges adjacent to a constant fraction of nodes in the graph are unobserved. In both of these models, we train our RL policy on the observed network and measure the performance of the learned policy on the actual network. We call these models edge and node removal, respectively. 

We point out that in the first model some of the properties of the original graph like the nodes with maximum degree are preserved~\cite{dubois2012effect}. In the second model, many of the properties of the original network like centrality are likely to be changed~\cite{smith2013structural}.

Although our approach is the only learning algorithm that can benefit from different training sub-graphs, to make fair comparison, we train our RL policy on 
%have each algorithm calculate or train its policy based on 
a single sub-graph. This is corresponding to the real world scenario where partial contact information is missing without being noticed. The results are summarized in Figures~\ref{uncertain_edge}~and~\ref{uncertain_node}, respectively. Although the performance of our approach decays as the uncertainty increases, it still outperforms all the other baselines. Again, we show the result of face-to-face network as an example due to space limit. The results of the other networks have similar trends and can be found in Appendix~\ref{app:exp}.
%\sj{need to summarize the results, add some insight and also mention why we are focusing on specific networks and maybe put the results for other networks into appendix}

\begin{figure}[ht!]
\includegraphics[width=0.4\textwidth,keepaspectratio]{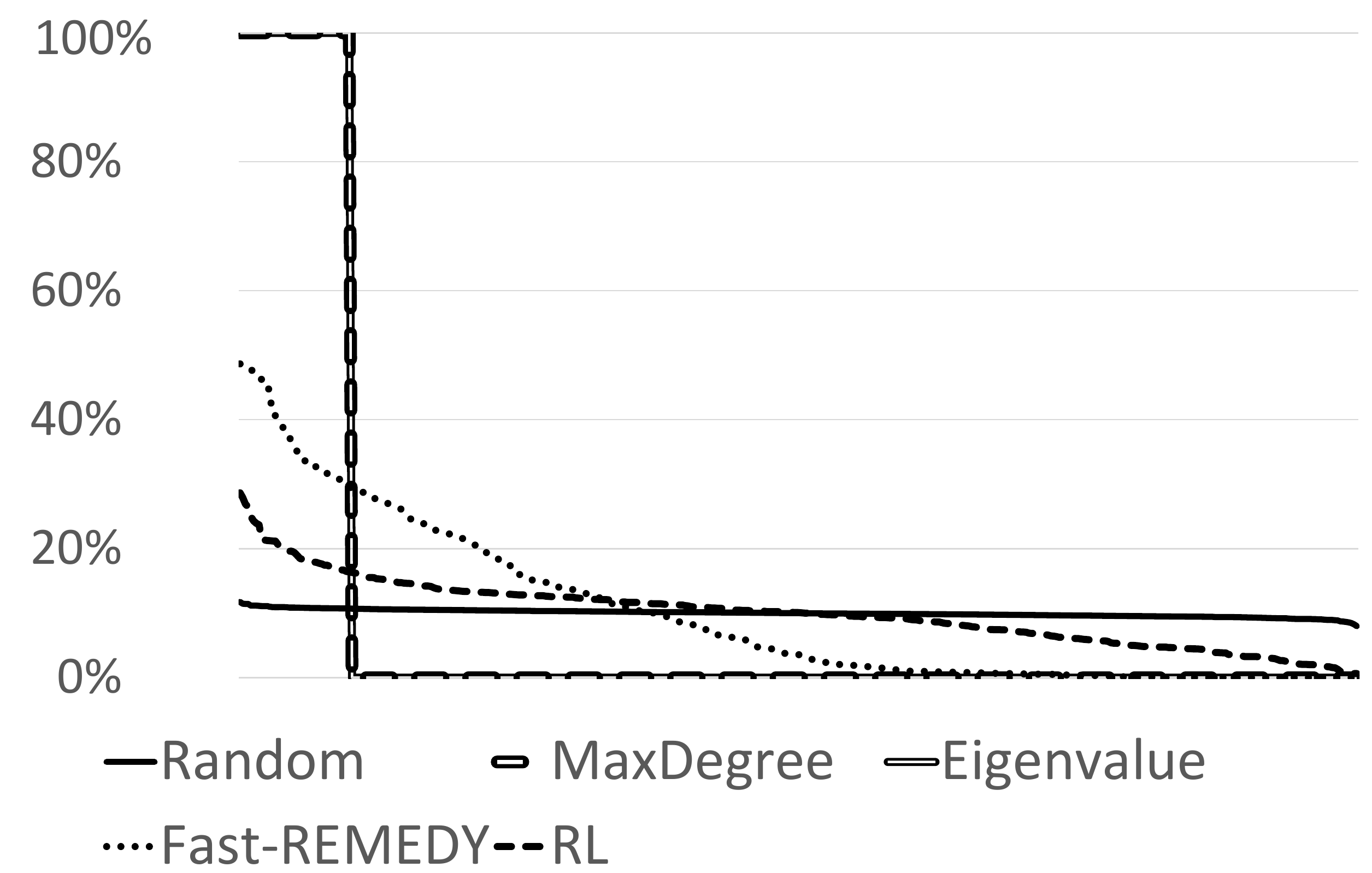}
\centering
\caption{The node picking frequency distribution for each algorithm, sorted from high to low.}\label{node_freq}
\end{figure}

\subsection{Policy Analysis}
\begin{figure}[ht!]
\centering
\null\hfill
\subfloat[Average Degree]{%
  \includegraphics[width=0.24\textwidth,keepaspectratio]{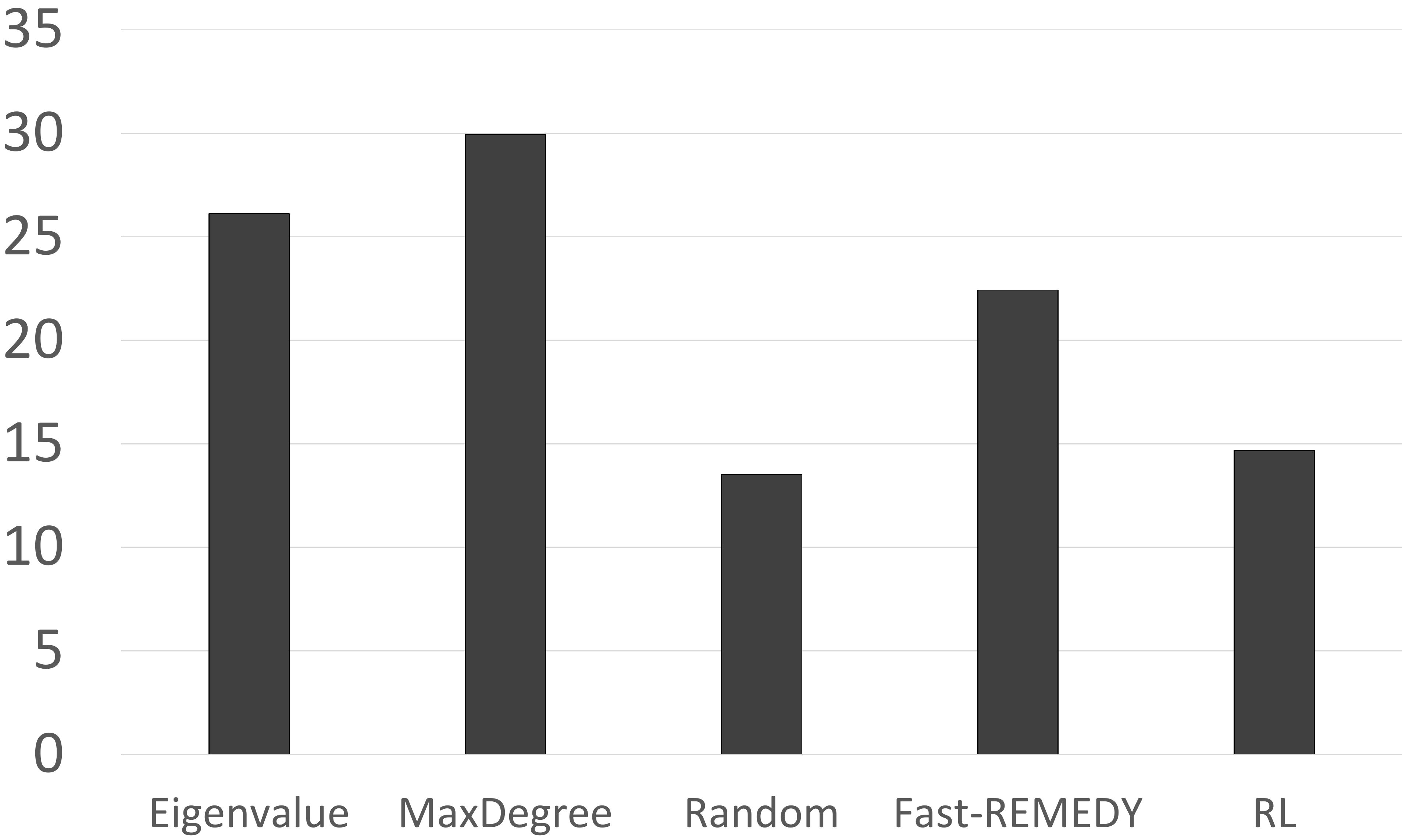}%
}
\subfloat[Average Betweeness]{%
  \includegraphics[width=0.24\textwidth,keepaspectratio]{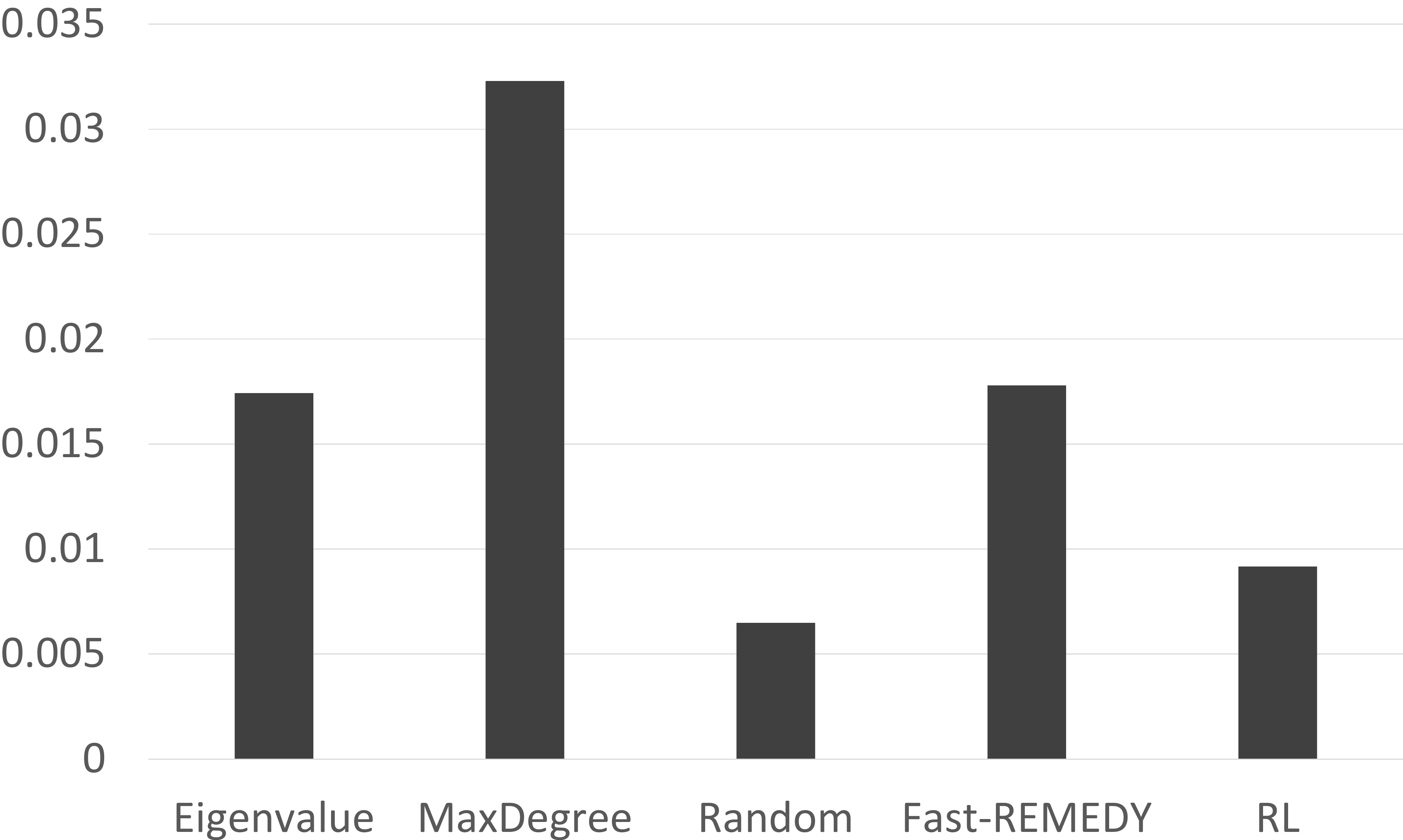}%
}\hfill\null
\centering
\vspace*{-3.0ex}
\caption{Average degree \& betweeness of the nodes picked.}\label{properties}
\end{figure}

To gain insight on the patterns of how different approaches select nodes, we study the frequency in which each of the nodes is selected. The results are summarized in Figure~\ref{node_freq} for the Face-to-face network. Similarly, results for other networks are in Appendix~\ref{app:exp}. Each point in x-axis represents a node, sorted by the frequency of being picked by the corresponding algorithm. The y-axis represents the frequency a certain node is picked by the algorithm. We sort all the algorithm's node picking frequency in order to show their distribution. In this figure, \textit{Random} is the fairest algorithm as it picks each node with equal frequency, whereas \textit{MaxDegree} and \textit{Eigenvalue} always pick the same set of nodes as we have a static network. \textit{Fast-REMEDY} selects the most frequently picked nodes half of the times while almost never picks $40\%$ of the nodes. Our RL approach does not pick the structurally important nodes as often as \textit{Fast-REMEDY} does, which is shown in figure~\ref{properties}. It is less structure dependent and tends to select a larger variety of nodes.
By taking future actions into account, it depends more on the observation information and has a surprising side effect that outputs a fairer policy that gives more nodes chances to be screened. %\sj{we need to describe the differences between figures 6 and 7. also need to add more insights}

\vspace*{-3.0ex}
\vspace*{-3.0ex}

\section{Conclusion}
We make the first attempt at addressing the multi-round active screening problem using reinforcement learning. We formulate the problem as a MDP with high dimensional state and action spaces, which cannot be efficiently solved using classical RL algorithms like DQN. We then design several innovative adaptations to vanilla DQN, including GCN-based value function approximator that exploits the correlations of nodes, a primary-secondary agents framework that decomposes the combinatorial action selection in each time period into a sub-sequence of node selection, and a curriculum learning component that addresses the sparseness of reward for the secondary agents. Empirical results show that in terms of solution quality, our approach outperforms the state-of-the-art \textit{Fast-REMEDY} by a margin of $9\% - 33\%$, and works better than baselines even with network structure uncertainty. In the largest network we experimented which 1899 nodes, our approach is able to scale up to a planning horizon 10 times that in state-of-the-art approach \textit{Full-REMEDY}. Interestingly, policy analysis results show that compared with most baselines (except for \textit{Random}), our approach is fairer in the sense that it tends to spread the screening across different nodes. For future work, we plan to incorporate uncertainty on the graph structure in training our RL algorithms and further improve the robustness of our approach.
%Our approach can be potentially extended to the active screening problem with structure uncertainty using different graph embedding approaches for future work \sj{you can say as future work we can incorporate uncertainty ...}, or apply to other sequential combinatorial problem.
%\sj{I have a hard time following this section}

%\textbf{Acknowledgments}: Chen and Jabbari was supported by the Center for Research on Computation and Society. This work was supported by the Army Research Office (MURI W911NF1810208).

\begin{acks}
Chen and Jabbari were supported by the Center for Research on Computation and Society. This work was supported by the Army Research Office (MURI W911NF1810208). This research was also supported by NSF CCF-1522054.
\end{acks}

%%%%%%%%%%%%%%%%%%%%%%%%%%%%%%%%%%%%%%%%%%%%%%%%%%%%%%%%%%%%%%%%%%%%%%%%

%%% The next two lines define, first, the bibliography style to be 
%%% applied, and, second, the bibliography file to be used.
\newpage\clearpage
\bibliographystyle{ACM-Reference-Format} 
\bibliography{main}

%%%%%%%%%%%%%%%%%%%%%%%%%%%%%%%%%%%%%%%%%%%%%%%%%%%%%%%%%%%%%%%%%%%%%%%%
%\clearpage
%\appendix
%\input{Appendix}

\end{document}